\documentclass[conference, letterpaper]{IEEEtran}

\usepackage[utf8]{inputenc}
\usepackage[cmex10]{amsmath}
\usepackage[pdftex]{graphicx}
\usepackage{titlesec}
\usepackage{xcolor}
\usepackage[caption=false, font=footnotesize]{subfig}
\usepackage{adjustbox}
\usepackage{booktabs}
\usepackage{algorithmic}
\usepackage[ruled, vlined]{algorithm2e}
\usepackage{tikz}
\usetikzlibrary{arrows.meta, positioning, shapes.geometric}
\usepackage{textcomp}
\usepackage{enumitem}
\usepackage{listings}
\usepackage{amssymb}
\usepackage{mathptmx}
\usepackage{amsfonts}
\usepackage{fancyhdr}
\usepackage{url}
\usepackage{xurl}
\usepackage{hyperref}
\hypersetup{
    colorlinks=true,
    linkcolor=red,
    citecolor=blue,
    urlcolor=black,
    filecolor=black,
    linkbordercolor={0 0 0},
    pdfborder={0 0 0}
}
\usepackage{float}

\renewcommand{\thispagestyle}[2]{}

\newcommand{\Labels}{\mathcal{Y}}
\newcommand{\Context}{\textsf{Context}}

\newcommand{\Act}{\textsf{Activity}}

\definecolor{accent}{HTML}{374785}
\definecolor{ritorange}{HTML}{F76902}

\SetInd{0.35em}{0.6em}
\SetAlgoInsideSkip{smallskip}
\DontPrintSemicolon

\fancypagestyle{plain}{
    \fancyhead{}
    \fancyhead[C]{first page center header}
    \fancyfoot{}
    \fancyfoot[C]{first page center footer}
}
\newcommand\tab[1][0.5cm]{\hspace*{#1}}

\title{Visual Timelines of Police Encounters in Body-Worn Camera Footage: \\
Operational Context and Activity Cataloging for Training and Analysis in OpenBWC}
\author{
    \IEEEauthorblockA{
        \begin{tabular}{cc}
            \begin{tabular}{c}
                Angela Srbinovska \\
                Department of Computer Science \\
                Rochester Institute of Technology \\
                Rochester, NY, USA \\
                \href{mailto:as2179@rit.edu}{as2179@rit.edu}
            \end{tabular}
            &
            \begin{tabular}{c}
                Christopher Homan \\
                Department of Computer Science \\
                Rochester Institute of Technology \\
                Rochester, NY, USA \\
                \href{mailto:cmhvcs@rit.edu}{cmhvcs@rit.edu}
            \end{tabular}
            \\ [-0.05cm]
            \multicolumn{2}{c}{\vspace{0.10cm}} \\
            \begin{tabular}{c}
                Adrian Martin \\
                Office of Business Intelligence \\
                Rochester Police Department \\
                Rochester, NY, USA \\
                \href{mailto:Adrian.Martin@CityofRochester.gov}{Adrian.Martin@CityofRochester.gov}
            \end{tabular}
            &
            \begin{tabular}{c}
                Ernest Fokou\'e \\
                School of Mathematics and Statistics \\
                Rochester Institute of Technology \\
                Rochester, NY, USA \\
                \href{mailto:epfeqa@rit.edu}{epfeqa@rit.edu}
            \end{tabular}
        \end{tabular}
    }
}

\begin{document}

\maketitle

% ---------------------------------------------------------------------------------------------------
%     ABSTRACT
% ---------------------------------------------------------------------------------------------------

\begin{abstract}
    Law enforcement agencies are accumulating vast amounts of body-worn camera (BWC) footage. However, this remains operationally opaque. That is, analysts and trainers still have to invest considerable time watching full-length videos to pinpoint the start of key encounters and identify the points where activity shifts to something more physically intense. We present an approach to process BWC video into a time-aligned sequence of fixed-length $\textbf{10}$-second windows, processed and labeled using a privacy-conscious protocol. Each window is labeled with two dimensions of information: (i) the operational context of the window and (ii) the level of motion intensity within the window, with low-evidence labels for windows for which insufficient evidence exists due to darkness, blur or occlusion. We train models to classify windows based on these two axes using frames sampled from each window encoded using CLIP model and aggregated into a window-level representation. We extract dense optical flow statistics for each window to capture motion intensity. On test windows the best context model achieves \textbf{$\textbf{78.75\%}$} accuracy, and the best-accuracy activity model achieves \textbf{$\textbf{88.33\%}$}. We also included integrity audits to show the results and how the visual timeline representations support faster incident review and make the officer training workflow more practical.
\end{abstract}


% ---------------------------------------------------------------------------------------------------
%     INTRODUCTION
% ---------------------------------------------------------------------------------------------------

\section{Introduction}\label{sec:intro}

\noindent Police departments have adopted body-worn cameras (BWCs) more over the past decade mainly as a way to capture video of officer-community interactions. As these systems have become more widespread, police agencies now maintain growing archives of BWC video, where individual events can extend from tens of minutes to over an hour. Even when a reviewer has a clear sense of the kind of event they are looking for, the standard approach still often requires watching footage in real time. Reviewers may be trying to answer questions such as:
\begin{quote}
    \emph{``When does the officer first enter a residence and when does a foot pursuit begin?''}
\end{quote}
\begin{quote}
    \emph{``How do police encounters change over time?''}
\end{quote}
\begin{quote}
    \emph{``Which segments show high activity motion?''}
\end{quote}
\tab For these reasons, we design a time-aligned structure that is easy to audit so that we can do systematic reviews on a large scale. In particular, we model two signals:
\begin{enumerate}[noitemsep]
    \item \textbf{Operational context:} a coarse scene state capturing the encounter setting, which we derive from appearance and layout statistics. We model context as a discrete label that typically evolves in long temporal runs.

    \item \textbf{Motion intensity:} a window-level kinematic regime summarizing the distribution of apparent motion over time. We summarize motion intensity using interpretable motion statistics.
\end{enumerate}
\tab We develop this project within the OpenBWC\footnote{OpenBWC is an on-site, access-controlled multi-sensory system codesigned with Rochester Police Department (RPD) partners to assist in the privacy-governed research and reviewing of BWC incidents. \texttt{\url{https://openbwc.org/}}} collaboration that includes the Rochester Institute of Technology (RIT), the Rochester Police Department (RPD), and criminology partners from the University at Albany. Existing OpenBWC components focus on audio and text workflows---transcription, speaker diarization, and human-in-the-loop correction---producing time-aligned transcripts for downstream analysis. These capabilities support queries over \emph{what was said} and \emph{when it was said}~\cite{srbinovska2025openbwc}. \\
\tab However, the transcripts are just the spoken language, and as such, cannot account for the shared situational context or the physical dynamics of an encounter. For example during a typical interaction the officer and the other person treat the premises and the ongoing actions as common ground, resulting in speakers not frequently articulating the particulars. Consequently, the video record supplies critical review-relevant information that is not generally included in the transcript, such as the officer being in a patrol vehicle, on a street, approaching a doorway or in a house. The paper introduces a visual understanding layer and provides additional contextual and activity information in the form of time-stamped labels, which leads to the  following research questions:
\begin{enumerate}[label={}, noitemsep]
    \item \textbf{RQ$\textbf{1.}$} Do $10$-second video windows mostly show stable parts of an encounter?

    \item \textbf{RQ$\textbf{2.}$} How much of the footage is difficult to label due to darkness and does that problem happen mostly in specific incidents?

    \item \textbf{RQ$\textbf{3.}$} Do the patterns of physical activity look different depending on the operational context?

    \item \textbf{RQ$\textbf{4.}$} Using minimal baseline models, can we predict the operational context more reliably than the level of motion intensity?

    \item \textbf{RQ$\textbf{5.}$} When the model gives a low-confidence prediction, does that tend to align with the windows that human annotators also found difficult?
\end{enumerate}
\tab For RQ$1$--RQ$3$ we perform dataset audits (Section~\ref{sec:data:quality}). We answer RQ$4$ and RQ$5$ with predictive evaluation and error analysis (Section~\ref{sec:evaluation}). \\
\tab In summary we make the following contributions:
\begin{itemize}[noitemsep]
    \item \textbf{a reproducible visual timeline method} that keeps privacy intact (no face/person recognition) and gives a clear, time-stamped, auditable index.
    
    \item \textbf{a labeling scheme} that is based on what can be seen in real BWC footage conditions.
    
    \item \textbf{simple, transparent baseline classification models} that require minimal supervised training.
    
    \item \textbf{an evaluation approach} combining class-wise metrics and confusion matrices.
\end{itemize}
\tab This paper is structured as follows. Section~\ref{sec:related_Work} reviews related work in BWC research. Section~\ref{sec:data} describes the data corpus, windowing process and annotation protocol. Section~\ref{sec:methods} presents the modeling approach. Section~\ref{sec:evaluation} shows the experimental setup. Section~\ref{sec:discussion} concerns limitations. Section~\ref{sec:conclusion} concludes and Section~\ref{sec:future_work} lists future directions.

% ---------------------------------------------------------------------------------------------------
%     RELATED WORK
% ---------------------------------------------------------------------------------------------------

\section{Related Work}\label{sec:related_Work}

\noindent Prior work on police BWC footage comes from fields such as criminology, public policy and computer science. For this paper, we focus on the following related areas: outcomes, structured coding, review systems, and egocentric video understanding. \\
\tab Early research on police BWCs looked into why departments decided to start using them, what kinds of problems arose during the process and the bigger debates about making policing more open and holding officers accountable~\cite{white2014police}. Studies moved toward trying to measure actual effects, such as whether activated cameras led to fewer instances where officers used force, or fewer complaints against the police~\cite{ariel2015wearing}. \\
\tab BWCs became material that could be coded, compared and studied across incidents. In that setting, video-based systematic social observation~\cite{mccluskey2023vsso,terrill2022police} and related coding approaches showed that encounters could be analyzed in a consistent way across many cases~\cite{worden2025sso_bwc}. Holladay and Makin offer one example through their work on incivility in one-on-one encounters drawn from archival BWC footage~\cite{holladay2021incivility_bwc}. Their study shows that structured behavioral coding is useful and that it still needs a lot of human judgment. A related work has grown around transcripts generated from BWC audio. So researchers have used these transcripts to check officer communication mostly during traffic stops~\cite{camp_pnas_2024} and to apply natural language processing methods to support officer training and review of police-citizen interactions~\cite{camp_voigt_2024}. \\
\tab Watching manually hours of BWC footage can be time-consuming. Annotation tools were created out of necessity to transform unstructured video into usable datasets for testing and training as well as future modeling. More recent developments have included the release of web-based platforms specifically created for labeling police BWC footage. CVAT-BWV is an open-source extension of the Computer Vision Annotation Tool (CVAT)~\cite{hejabi2024vat_bwv}. CVAT-BWV is built to be deployed locally to ensure that departments can host the program on their own computers rather than uploading media to a third party cloud. The platform includes features such as automatic speech recognition to automatically generate transcripts, speaker diarization, object detection and face related functionalities that support privacy. These tools allow departments to reduce time input for labeling videos, ensure better consistency between labels and allow for more annotators. Work with real-world BWC benchmark datasets shows that generalization remains a major challenge for machine learning models. Results often weaken when models are tested on footage from a different department, a new camera model or another mounting angle~\cite{hans2025faleb_dataset}. \\
\tab Existing research into computer vision, described this as a ``high-impact, but technically challenging space,'' which focused on the discrepancy between what the agency needs and what the unedited video from BWCs can provide~\cite{corso2016video_analysis_bwc}. More recent research focused on the development of learning-based approaches to BWC video recognition. For example semi-supervised first-person activity recognition explores how to exploit unlabeled BWC data to reduce annotation demand and stabilize learning when labeled examples are scarce~\cite{chen2019semi_supervised_bwc}. A complementary work takes the problem less in terms of ``which action is occurring'' and more in terms of ``the degree of movement exhibited by the camera wearer.'' Research into ego-motion classification suggests that the dynamics of the motion can carry a significant portion of the information that is useful such as walking, running, and fast acceleration even when the visual information is degraded~\cite{2018ego_motion_bwc}. Researchers have long made use of representations of motion, such as optical flow, to segment or characterize the different phases of the video, as they summarize the information over the course of the video~\cite{poleg2014egocentric}. \\
\tab The recent advances in video models and vision-language pretrained models may also open the door for alternative solutions for video-based retrieval and labeling that do not rely on task-specific models~\cite{madan2024foundation_models,wang2024internvideo2,romero2023zelda}. In particular, the interest in vision-language models such as CLIP\footnote{CLIP = Contrastive Language-Image Pretraining} has revived interest in zero-shot classification schemes~\cite{radford2021learning,cherti2022reproducible}. Nevertheless, in order to utilize these models in the context of BWC, there are also additional factors including data-use policies for agencies, redactions that may obscure the information at the level of persons, and the need to simplify model outputs existing schemas, rather than semantic knowledge graphs.

% ---------------------------------------------------------------------------------------------------
%     DATA
% ---------------------------------------------------------------------------------------------------

\section{Data}\label{sec:data}

\noindent This section documents the video corpus and data artifacts used for this research, as well as the deterministic windowing, annotation taxonomy and audits used to verify integrity and characterize uncertainty.

\subsection{Video Corpus}

\noindent This work employs a subset of RPD BWC videos maintained and processed by the OpenBWC project team through RIT Research Computing (RC)~\cite{rit_research}\footnote{RIT RC is the university’s centrally managed research computing environment that provides high-performance computing and research data storage services under institutional access controls.}. This subset includes various BWC video camera types, which provide a variety of field-of-view perspectives, lens types, image stabilization modes and video recording settings. Lighting conditions also show considerable diversity across the videos. Due to the evidentiary nature of the videos, the MP$4$ videos are kept unchanged. We provide a reference to the original incident video and the interval at which the incident happened. By ``curated,'' we mean that we selected a smaller subset from the full set while still preserving a range of recording conditions. \\
\tab Before windowing, annotating or training models we automatically performed a readiness assessment for each video. This process for a video includes: (i) opening the file and parsing stream metadata, (ii) decoding the first frame, excluding metadata parsing failures and (iii) extracting basic information from the video file. Table~\ref{tab:corpus-summary} shows the readiness results and dataset statistics for the subset used in this study.
\begin{table}[h]
    \centering
    \caption{Scope of the curated RPD BWC corpus used in this study.}
    \label{tab:corpus-summary}
    \footnotesize
    \begin{tabular*}{\columnwidth}{@{\extracolsep{\fill}} l r @{}}
        \toprule
        \textbf{Measure} & \textbf{Value} \\
        \midrule
        Incident videos & $15$ \\
        Total footage (hr) & $2.19$ \\
        Video duration (sec) & min $233.43$,\; max $1131.93$ \\
        Frame rate (fps) & min $29.97$,\; max $30.00$ \\
        Total windows & $783$ \\
        \bottomrule
    \end{tabular*}
\end{table}

\noindent \tab Using the MP$4$ videos that we validated, we divide each incident video $v$ into non-overlapping windows of fixed length $L = 10$ seconds. This gives $N_v = \lfloor D / L \rfloor$ complete windows, where $D$ denotes the video duration. Each window is mapped to a timestamp interval $[t, t + L)$. We discard the remaining portion of the video, if any, so that every labeled window has the same $10$-second length.

\subsection{Annotation Taxonomy}

\noindent \tab We assign two labels to each $10$-second video window: a context label, which describes the type of scene, and an activity label, which describes the intensity of motion, as shown in Tables~\ref{tab:context-defs} and~\ref{tab:activity-defs}. The labels are descriptive and only use information visible within the window. Labels do not include any information about the identities of persons seen in the video, or make judgments about behavior or performance. When visual evidence is limited, annotators use low-evidence labels. These labels are a core design choice of this work because they avoid forced guessing, reduce noisy supervision, and prevent overconfident labeling.
\begin{table}[h]
    \centering
    \caption{Definitions of the operational context labels.}
    \label{tab:context-defs}
    \begin{adjustbox}{max width=\columnwidth}
        \begin{tabular}{l p{0.69\columnwidth}}
            \toprule
            \textbf{Label} & \textbf{Definition} \\
            \midrule
            \texttt{PATROL\_VEHICLE} &
            Distinct vehicle interior structure (dashboard, windshield framing, steering wheel), frequent reflections and vehicle-interior viewpoint cues typical of driving. \\
            \texttt{OUTDOOR} &
            Open environment signals such as streets, sidewalks, sky, building exteriors, parked vehicles and large depth range. \\
            \texttt{INDOOR} &
            Clear interior signals such as walls, ceilings, doorframes, hallways, indoor lighting and enclosed surfaces. \\
            \texttt{LOW\_VIS} &
            Insufficient visual evidence due to darkness, blur, occlusion (hand/clothing), and camera-down. This label represents uncertainty. \\
            \bottomrule
        \end{tabular}
    \end{adjustbox}
\end{table}
\begin{table}[h]
    \centering
    \caption{Definitions of the activity labels.}
    \label{tab:activity-defs}
    \begin{adjustbox}{max width=\columnwidth}
        \begin{tabular}{l p{0.69\columnwidth}}
            \toprule
            \textbf{Label} & \textbf{Definition} \\
            \midrule
            \texttt{ROUTINE} &
            Controlled motion, low-to-moderate motion magnitude and stable window-level behavior. \\
            \texttt{FOOT\_PURSUIT} &
            Sustained fast forward motion, large motion magnitudes with consistent direction across the window. \\
            \texttt{HIGH\_ACTIVITY} &
            Irregular high motion and magnitude with low directional coherence and high variance. \\
            \texttt{UNKNOWN} &
            Window lacks reliable visual evidence to assess motion level. \\
            \bottomrule
        \end{tabular}
    \end{adjustbox}
\end{table}

\noindent \tab Because of the labor-intensive nature of manual labeling, we only sample a subset of the window inventory for labeling purposes. Let us denote the set of all $10$-second windows generated from the corpus as $\mathcal{W}$, and let us denote the subset of these windows that are labeled as $\mathcal{W}_L \subset \mathcal{W}$, where $|\mathcal{W}_L| = 428$. Sampling is done separately for each video. From the ordered set of windows based on \texttt{start\_time}, we select $k_{\text{cov}}$ windows (number of windows that should span the entire incident timeline) that are approximately evenly spaced, then select $k_{\text{rand}}$ windows uniformly at random from the remaining windows using a constant seed for reproducibility. The video quota for sampling is $k_{\text{cov}} + k_{\text{rand}}$ (fixed within a run), up to the total number of windows available in the video. \\ \\
\noindent \textbf{Dominant-time rule.} In cases where there are multiple labels, the label that appears the most in the given window of $10$ seconds is assigned. To differentiate between cases of clear supervision and boundary cases, the \texttt{context\_transition} and \texttt{activity\_transition} flags are also recorded. These flags are assigned the value of $1$ if a change in context and/or activity transpires in the window of $10$ seconds.

\subsection{Dataset auditing}\label{sec:data:quality}

\noindent Before any modeling, a series of automated dataset audits are performed to check the structural correctness of the dataset and understand the uncertainty and temporal behaviors within the annotated window. \\ \\
\noindent \textbf{Dataset quality checks.} We verified that the dataset had the following properties: that it contains $428$ unique window keys, and has zero missing labels and out-of-vocabulary values. In addition, we ensured that all window keys point to valid entries. \\ \\
\noindent \textbf{Label distributions.} We summarize the label distributions to show class imbalance and the share of low-evidence windows. The results in Table~\ref{tab:label_dist} show that \texttt{OUTDOOR} is the context we see often, making up $48.36\%$ of the time, and \texttt{ROUTINE} is the activity that happens most, which is $80.84\%$. When it comes to uncertainty, $19.16\%$ of labeled windows are classified as \texttt{LOW\_VIS}. This means about one out of every five windows does not give us information to figure out the context. On the other hand, we do not see \texttt{UNKNOWN} activity very often only $1.17\%$ of the time. This suggests that we can usually tell how intense the motion is, even if the window is not ideal. In our sample every time we saw \texttt{UNKNOWN} activity it was due to context. Still, we do not treat this as a general rule. In a larger and more varied dataset, other conditions such as glare, overexposure, strong motion blur or upward camera angles could also make activity difficult to label.
\begin{table}[h]
    \centering
    \caption{Class distribution in the labeled subset.}
    \label{tab:label_dist}
    \footnotesize
    \setlength{\tabcolsep}{6pt}
    \begin{tabular*}{\columnwidth}{@{\extracolsep{\fill}} l r r @{}}
        \toprule
        \textbf{Context} & \textbf{Count} & \textbf{\%} \\
        \midrule
        \texttt{OUTDOOR} & $207$ & $48.36$ \\
        \texttt{LOW\_VIS} & $82$ & $19.16$ \\
        \texttt{PATROL\_VEHICLE} & $71$ & $16.59$ \\
        \texttt{INDOOR} & $68$ & $15.89$ \\
        \midrule
        \textbf{Activity} & \textbf{Count} & \textbf{\%} \\
        \midrule
        \texttt{ROUTINE} & $346$ & $80.84$ \\
        \texttt{HIGH\_ACTIVITY} & $53$ & $12.38$ \\
        \texttt{FOOT\_PURSUIT} & $24$ & $5.61$ \\
        \texttt{UNKNOWN} & $5$ & $1.17$ \\
        \bottomrule
    \end{tabular*}
\end{table}

\noindent \textbf{Context--activity relationship.} We next test whether context and activity labels are statistically independent. Table~\ref{tab:cond_rates} reports conditional rates suggesting a small-to-moderate dependence between context and activity. 
\begin{table}[h]
    \centering
    \caption{Conditional activity distribution given context labels.}
    \label{tab:cond_rates}
    \scriptsize
    \setlength{\tabcolsep}{3pt}
    \begin{tabular}{l r r r r}
        \toprule
        \textbf{} & \texttt{ROUTINE} & \texttt{HIGH\_ACTIVITY} & \texttt{FOOT\_PURSUIT} & \texttt{UNKNOWN} \\
        \midrule
        \texttt{INDOOR} & $0.868$ & $0.088$ & $0.044$ & $0.000$ \\
        \texttt{OUTDOOR} & $0.816$ & $0.082$ & $0.101$ & $0.000$ \\
        \texttt{PATROL\_VEHICLE} & $0.958$ & $0.042$ & $0.000$ & $0.000$ \\
        \texttt{LOW\_VIS} & $0.610$ & $0.329$ & $0.000$ & $0.061$ \\
        \bottomrule
    \end{tabular}
\end{table}

\noindent \tab In \texttt{OUTDOOR} windows \texttt{FOOT\_PURSUIT} occurs $10.1\%$ of the time. \texttt{PATROL\_VEHICLE} windows are predominantly \texttt{ROUTINE}. \texttt{UNKNOWN} activity occurs only when context is \texttt{LOW\_VIS} and within \texttt{LOW\_VIS} windows, \texttt{UNKNOWN} happens $6.1\%$ of the time.

\noindent \textbf{\\ Temporal stability.} In Table~\ref{tab:top_transitions} we examine how frequently the labels change between adjacent labeled windows. The overall context change rate is $\bar r(\text{context}) = 0.1356$, while the activity change rate is $\bar r(\text{activity}) = 0.1719$. This indicates that shifts in activity tend to occur somewhat more often than changes in scene.
\begin{table}[h]
    \centering
    \caption{Most frequent label transitions between adjacent windows.}
    \label{tab:top_transitions}
    \scriptsize
    \begin{tabular*}{\columnwidth}{@{\extracolsep{\fill}} l r l r @{}}
        \toprule
        \textbf{Context} & \textbf{Count} & \textbf{Activity} & \textbf{Count} \\
        \midrule
        OUT$\rightarrow$OUT & $178$ & R$\rightarrow$R & $306$ \\
        LV$\rightarrow$LV & $62$ & H$\rightarrow$R & $20$ \\
        OUT$\rightarrow$LV & $13$ & R$\rightarrow$H & $17$ \\
        \bottomrule
    \end{tabular*}
    
    \vspace{0.3em}
    \footnotesize \emph{Abbrev:} OUT=\texttt{OUTDOOR}, PV=\texttt{PATROL\_VEHICLE}, IN=\texttt{INDOOR}, LV=\texttt{LOW\_VIS}; R=\texttt{ROUTINE}, H=\texttt{HIGH\_ACTIVITY}, F=\texttt{FOOT\_PURSUIT}.
\end{table}

\begin{figure}[h]
    \centering
    \includegraphics[width=\columnwidth]{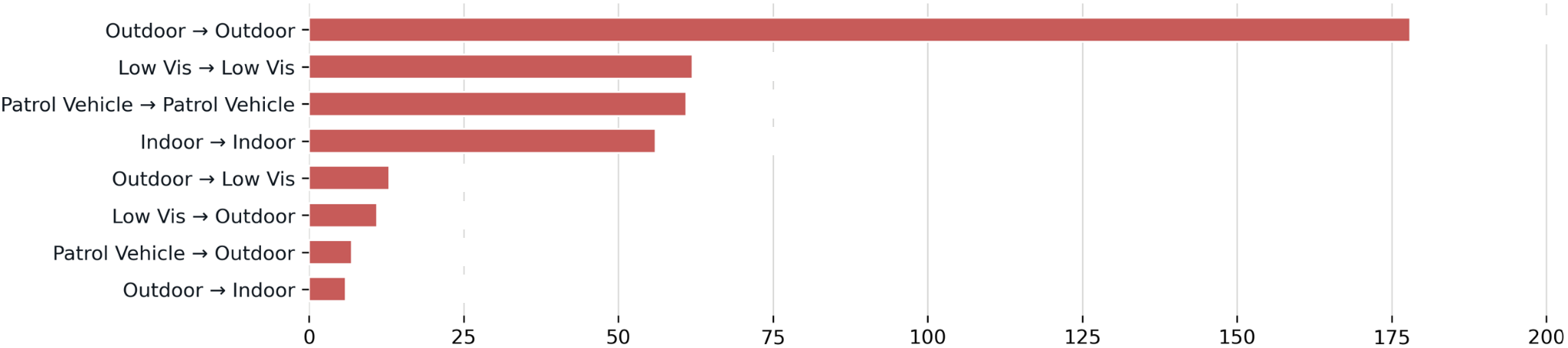}
    
    \vspace{0.5em}
    
    \includegraphics[width=\columnwidth]{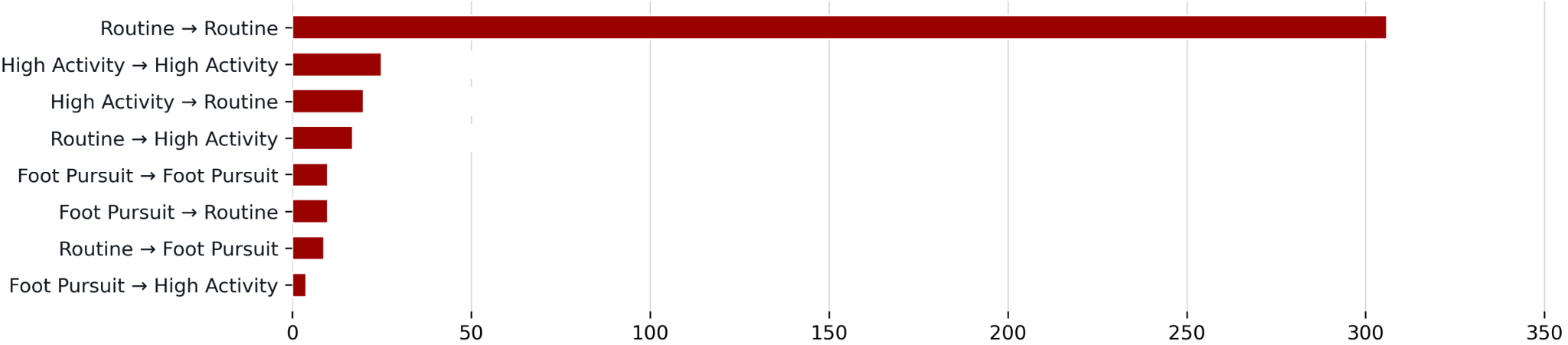}
    \caption{Most frequent transitions between adjacent labeled windows in the labeled subset. Top: context transitions. Bottom: activity transitions.}
    \label{fig:transition_counts}
\end{figure}

\noindent \tab Figure~\ref{fig:transition_counts} provides a more detailed view of adjacent-window transition patterns for both label spaces.

\begin{figure}[h]
    \centering
    \includegraphics[width=\columnwidth]{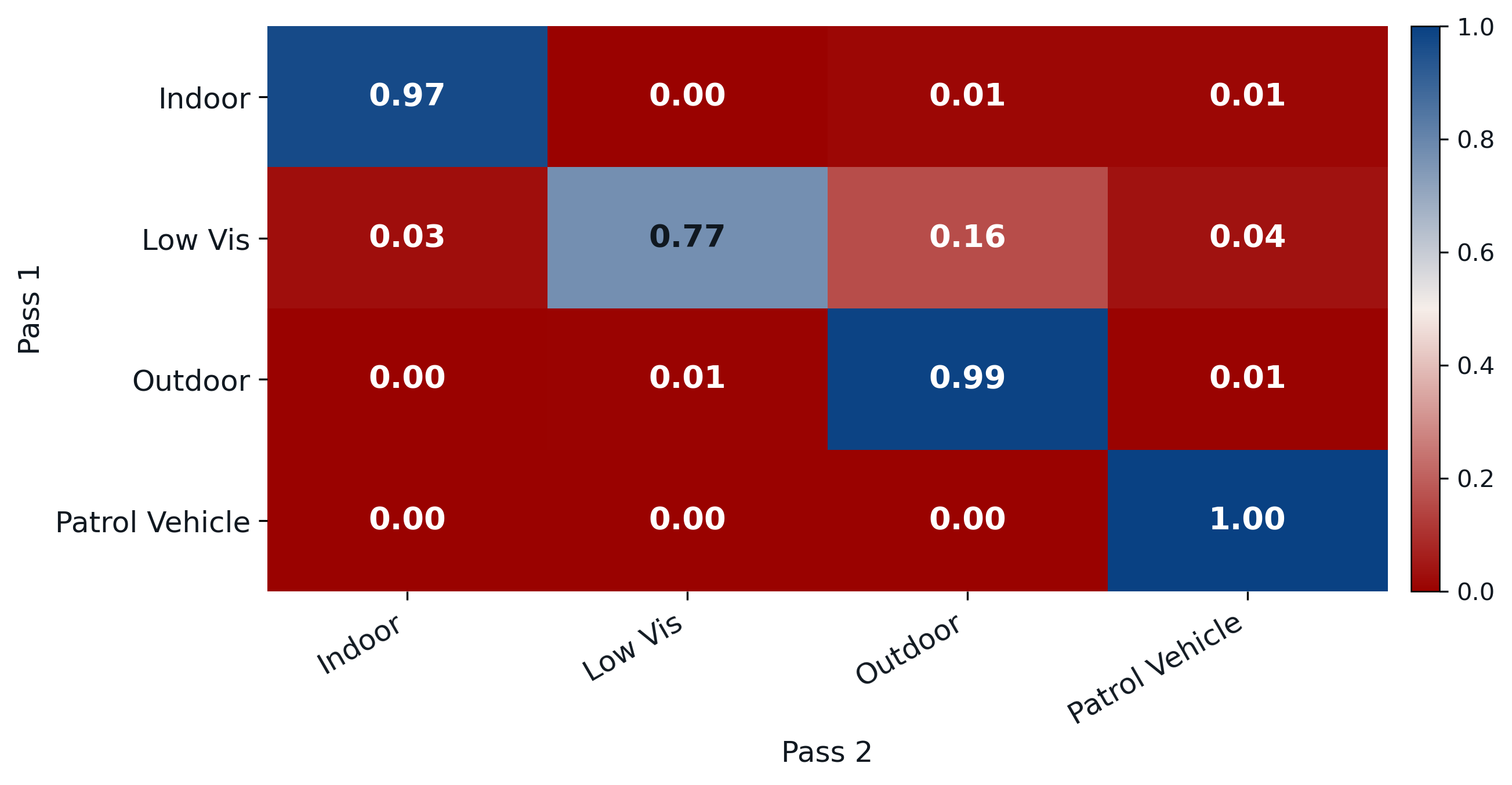}
    
    \vspace{0.5em}
    
    \includegraphics[width=\columnwidth]{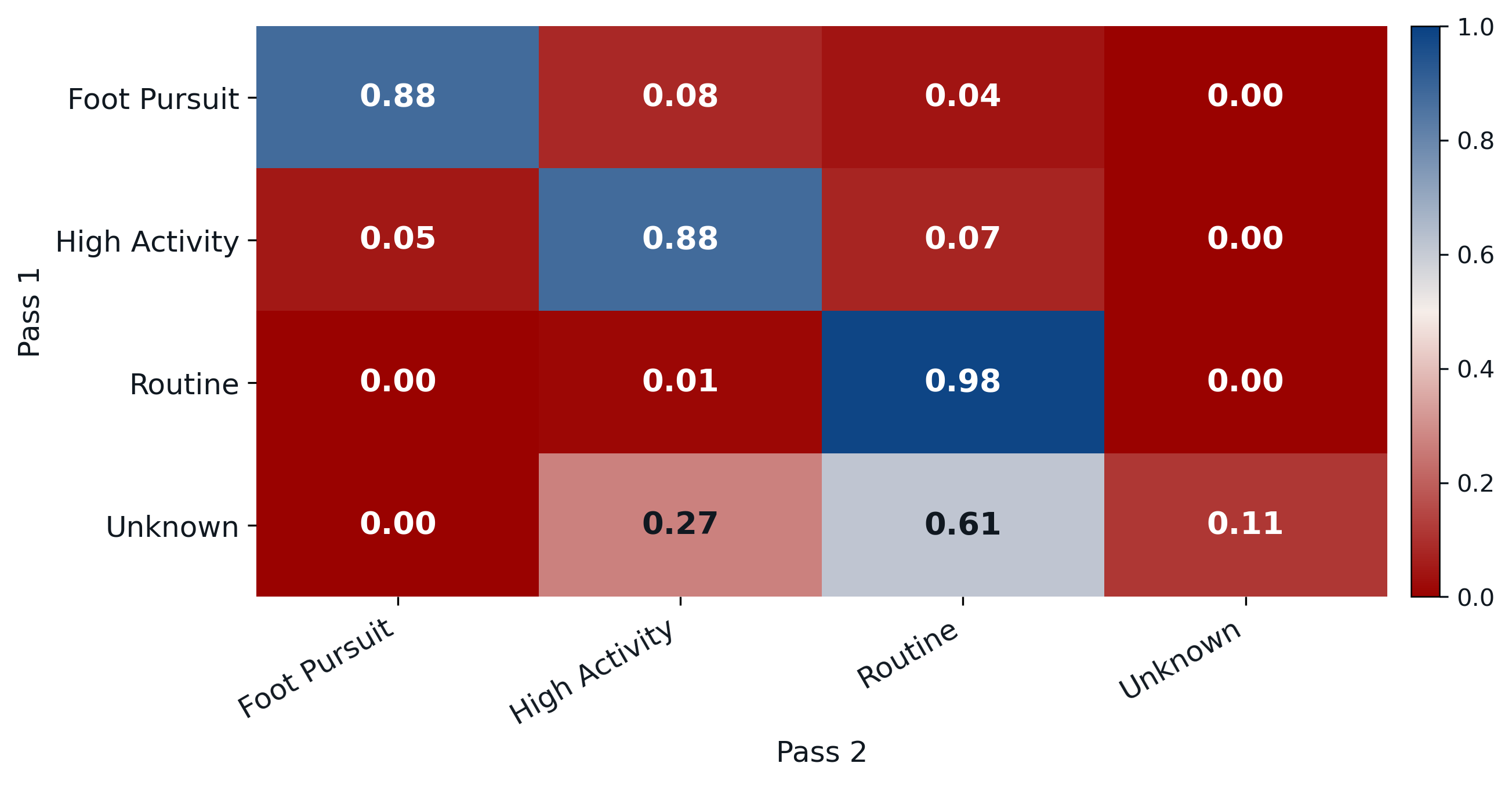}
    \caption{Row-normalized agreement matrices comparing human annotator labels from Pass $1$ and Pass $2$. Top: context agreement. Bottom: activity agreement.}
    \label{fig:agreement_heatmaps}
\end{figure}

\noindent \tab In the context labels, self-transitions are by far the most common. When the label does change, it is usually between just a handful of pairs, especially \texttt{OUTDOOR} flipping back and forth with \texttt{LOW\_VIS}. Overall, it shows that the scene labels tend to stay stable from one window to the next. The activity labels follow a similar pattern, with self-transitions dominating. \texttt{ROUTINE}$\rightarrow$\texttt{ROUTINE} is more frequent than anything else. That said, activity labels switch between neighboring windows more often than context does, particularly bouncing between \texttt{ROUTINE} and \texttt{HIGH\_ACTIVITY}. \\ \\
\noindent \textbf{Run-length behavior.} In addition to considering the rate at which the label of adjacent windows differs, it is also useful to consider how long any individual label remains visible. This provides a distinct axis along which to judge the temporal structure of the context and activity. There is clearly some persistence in the context. For \texttt{PATROL\_VEHICLE} the median run length was $4$ windows, with a mean of $7.35$ windows. For the three settings \texttt{OUTDOOR}, \texttt{INDOOR} and \texttt{LOW\_VIS}, the median run lengths were $3$ windows each, with mean run lengths of $4.15$, $4.69$ and $4.71$ windows, respectively. The persistence of an activity state is measured in terms of both the median length of the longest contiguous run of windows and the mean run length. For the most common activity state, the majority of the longest runs of contiguous windows that were analyzed consisted entirely of \texttt{ROUTINE} and had a length of $5$ windows. The mean run length for \texttt{ROUTINE} was $8.65$ windows. In contrast, the briefly mentioned \texttt{FOOT\_PURSUIT}, \texttt{HIGH\_ACTIVITY} and \texttt{UNKNOWN} each had a median run length of $1$ window and, accordingly, a mean run length of $1.71$, $1.89$ and $1.25$ windows, respectively. \\ \\
\noindent \textbf{Within-window transition burden.} We also looked at how often there was a shift inside a single $10$-second window. In the labeled data, $61$ out of $428$ windows (about $14.3\%$) had at least one internal transition. Most of them were activity-only shifts ($36$ windows or $59\%$) and the next most common were context-only shifts ($17$ windows or $27.9\%$). Just $8$ windows ($13.1\%$) had activity and context changes at the same time. These boundary problems usually happen because of changes in how much someone is moving, not because the scene itself changes suddenly. It also makes it clear why we should keep those explicit transition flags, so we can take those difficult windows out and look at them separately when we look at or review the results. \\ \\
\noindent \textbf{Annotator agreement.} To estimate labeling consistency, we compare two full labeling passes done by human annotators over the same labeled window set $\mathcal{W}_L$. We report exact-match agreement and Cohen's $\kappa$
\[
    \text{Acc}(y) \;=\; \frac{1}{|\mathcal{W}_L|}\sum_{w\in\mathcal{W}_L}\mathbb{I}\big[y^{(1)}(w) = y^{(2)}(w)\big] \ \text{and}
\]
\[
    \kappa \;=\; \frac{p_o - p_e}{1 - p_e},
\]
\noindent where $p_o$ is the observed agreement and $p_e$ is the chance of agreement, which we compute from the marginal label distributions. \\
\tab As shown in Figure~\ref{fig:agreement_heatmaps}, the context labels are more consistent across passes than the activity labels.
\noindent Activity agreement remains fairly strong, but it shows more variation across passes. Most of that movement comes from windows where the visual evidence is limited. In the context label space, agreement is strongly concentrated on the diagonal, especially for \texttt{PATROL\_VEHICLE}, \texttt{OUTDOOR} and \texttt{INDOOR}, while most disagreement involves \texttt{LOW\_VIS}, which is occasionally reassigned to neighboring scene categories. In the activity label space \texttt{ROUTINE} is the category that annotators agree on the most. Both \texttt{FOOT\_PURSUIT} and \texttt{HIGH\_ACTIVITY} show moderate agreement. The main source of disagreement is \texttt{UNKNOWN}. When labels are marked as \texttt{UNKNOWN} in Pass $1$, they are often changed in Pass $2$, most commonly to \texttt{ROUTINE} or \texttt{HIGH\_ACTIVITY}. \\
\tab These results confirm the fixed-window representation employed in this study. The timelines show significant temporal stability (\textbf{RQ$\textbf{1}$}). Uncertainty is constrained on challenging instances (\textbf{RQ$\textbf{2}$}). The observed connection between context and activity also shows that the representation captures important structure in police encounters (\textbf{RQ$\textbf{3}$}).

% ---------------------------------------------------------------------------------------------------
%     METHODS
% ---------------------------------------------------------------------------------------------------

\section{Methods}\label{sec:methods}

\subsection{Platform Description}

\begin{figure*}[h]
    \centering
    \includegraphics[width=\textwidth]{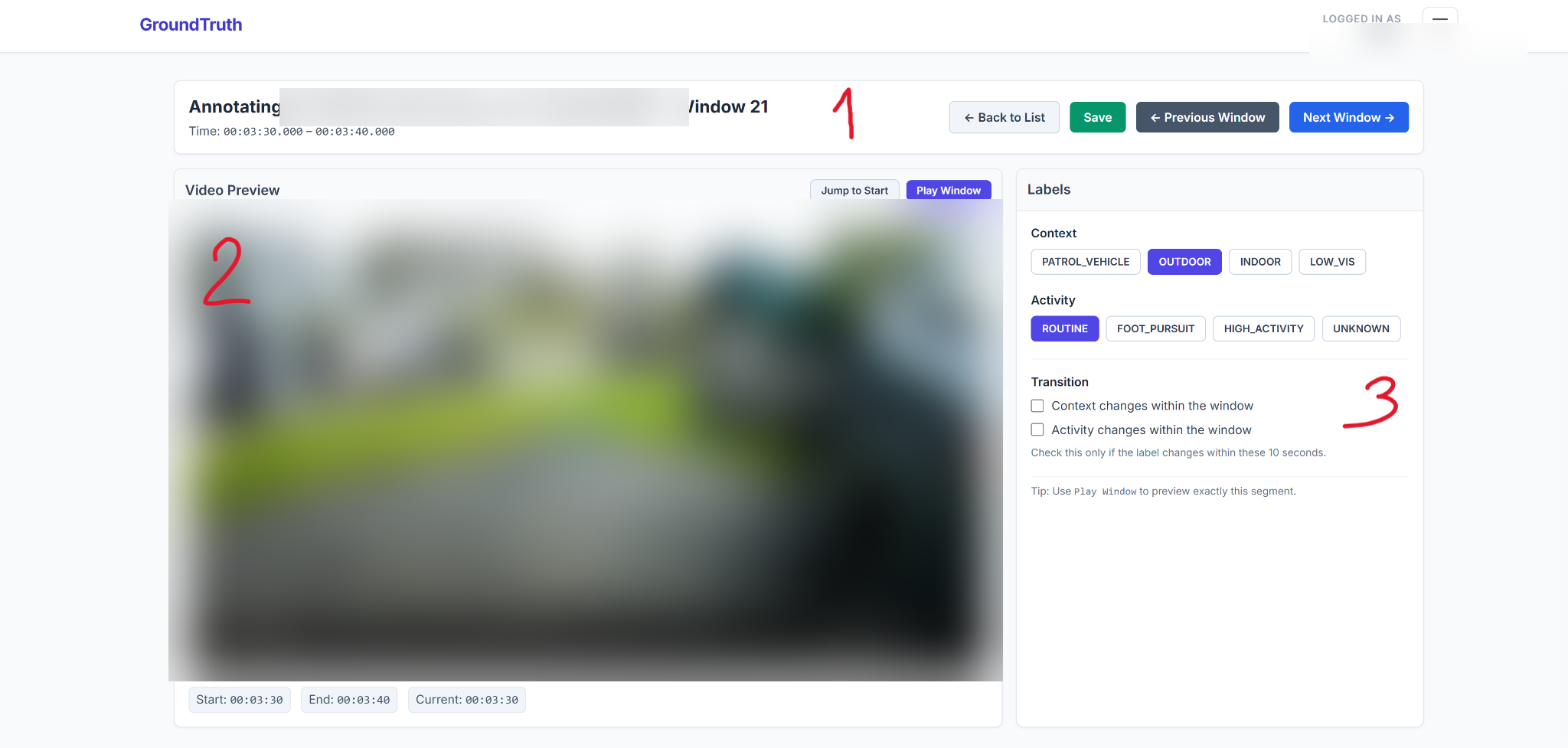}
    \caption{GroundTruth window-level annotation interface (video preview is intentionally blurred in this paper for privacy reasons). For every $10$-second window, a set of clipped timestamps is given requiring one context and one activity label from a fixed vocabulary. Annotators can optionally mark transitions happening within one window.}
    \label{fig:annotation-tool}
\end{figure*}
\noindent \textbf{\textit{GroundTruth}} is a web-based application that helps to label BWC videos for the OpenBWC timeline. GroundTruth is hosted in a secure environment. Saved labels are exportable for later review, training and evaluation. GroundTruth distinguishes itself from other labeling tools, since quick decisions on small parts of a video at a time are made handling scenarios with limited evidence where typically some boundary ambiguity exists, without allowing users to make wrong guesses. \\ \\
\noindent \textbf{User interface.} There are three main parts of the main annotation screen (see Figure~\ref{fig:annotation-tool}).
\begin{enumerate}[noitemsep]
    \item Window header: This area displays information about the segment to be annotated such as the incident ID, the specific window within the video that this segment is drawn from and the start and end timestamps of the annotated segment.

    \item Video panel: A preview player which displays a portion of the video the user is currently annotating. The user can navigate to the beginning of the currently selected segment and can play the same segment back again.

    \item Label panel: The annotator must select one context label and one activity label from a fixed vocabulary, and additionally can select a flag indicating whether or not there is any change within that segment.
\end{enumerate}

\subsection{Visual Timeline Pipeline}

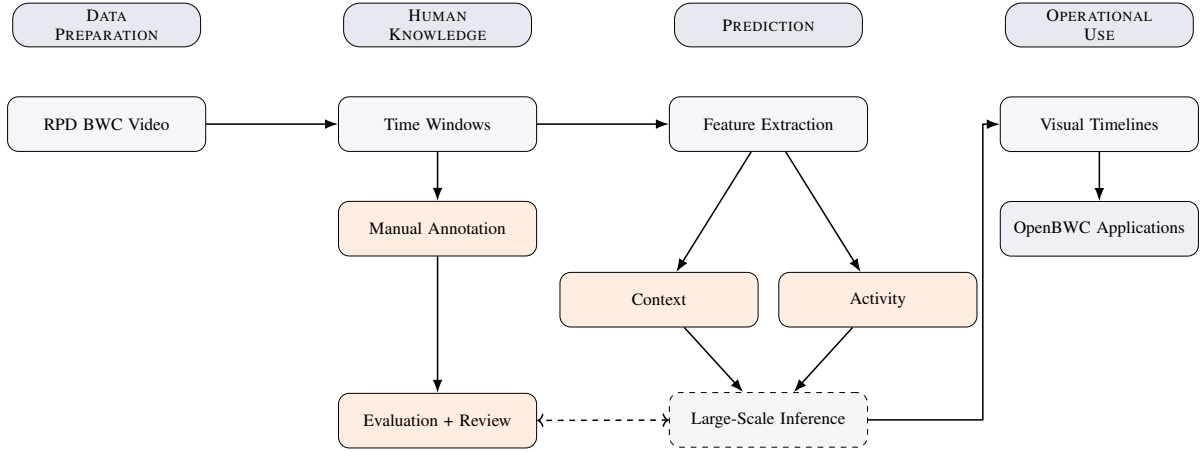
\begin{figure*}[h]
    \centering
    \begin{adjustbox}{max width=\textwidth, max height=0.25\textheight, center}
        \begin{tikzpicture}[
                node distance=12mm and 26mm,
                stage/.style={
                    draw, rounded corners=3mm,
                    align=center, font=\small\scshape,
                    minimum height=8mm, text width=3.2cm
                },
                box/.style={
                    draw, rounded corners=2mm,
                    align=center, font=\small,
                    inner sep=6pt,
                    minimum height=10mm, text width=3.2cm
                },
                smallbox/.style={
                    draw, rounded corners=2mm,
                    align=center, font=\footnotesize,
                    inner sep=5pt,
                    minimum height=8mm, text width=2.6cm
                },
                arrow/.style={-{Latex[length=2.4mm]}, thick}
            ]
            
            \node[stage, fill=accent!12] (stage1) {Data \\ Preparation};
            \node[stage, fill=accent!12, right=of stage1] (stage2) {Human \\ Knowledge};
            \node[stage, fill=accent!12, right=of stage2] (stage3) {Prediction};
            \node[stage, fill=accent!12, right=of stage3] (stage4) {Operational \\ Use};
            
            \node[box, below=9mm of stage1, fill=accent!5] (video) {RPD BWC Video};
            
            \node[box, below=9mm of stage3, fill=accent!5] (repr)
            {Feature Extraction};
            
            \node[box, below=9mm of stage4, fill=accent!5] (timeline)
            {Visual Timelines};
            
            \node[box, below=9mm of stage2, fill=accent!5] (windows) {Time Windows};
            
            \node[box, below=9mm of windows, fill=ritorange!12] (labels) {Manual Annotation};

            \node[box, below=25mm of labels, fill=ritorange!12] (eval)
            {Evaluation + Review};
            
            \node[box, below=22mm of repr, xshift=-20mm, fill=ritorange!12] (ctx_model)
            {Context};
            
            \node[box, below=22mm of repr, xshift=20mm, fill=ritorange!12] (mot_model)
            {Activity};
            
            \node[box, below=44mm of repr, dashed, fill=gray!8] (scale)
            {Large-Scale Inference};
            
            \node[box, below=9mm of timeline, fill=accent!8] (use)
            {OpenBWC Applications};
            
            \draw[arrow] (video) -- (windows);
            \draw[arrow] (windows) -- (labels);
            \draw[arrow] (windows) -- (repr);
            \draw[arrow] (repr) -- (ctx_model);
            \draw[arrow] (repr) -- (mot_model);
            \draw[arrow] (ctx_model) -- (scale);
            \draw[arrow] (mot_model) -- (scale);
            \draw[arrow] (timeline) -- (use);
            \draw[arrow] (labels) -- (eval);
            \draw[arrow, dashed, <->] (scale) -- (eval);
            \draw[arrow] (scale.east) -- ([xshift=1mm]mot_model.east |- scale.east) |- (timeline.west);
            \end{tikzpicture}
    \end{adjustbox}
    \caption{End-to-End Visual Analysis Pipeline.}
    \label{fig:pipeline}
\end{figure*}

\noindent In this work, we frame the task of visual timeline construction into two window-level supervised classification tasks. Namely, (i) operational context classification and (ii) activity (motion intensity) classification. \\
\tab Figure~\ref{fig:pipeline} shows the four stages of the workflow. In \textbf{$\textbf{(1)}$ Data Preparation}, BWC videos are partitioned into timestamped windows. In \textbf{$\textbf{(2)}$ Human Knowledge}, some of these windows are labeled by a human annotator using the controlled label space. In \textbf{$\textbf{(3)}$ Prediction}, time windows are analyzed. From a features perspective, lightweight classifiers are trained in a supervised fashion only using the labeled subset of time windows. The resulting models can then be used to predict the labels and the corresponding confidence scores for all windows in the video. In \textbf{$\textbf{(4)}$ Operational Use}, the predicted streams are used to generate unified timeline that can be consumed by OpenBWC or more general law enforcement applications. The dashed line in the evaluation loop indicates that the evaluation results can be used to adjust the thresholds and the sampling policies without modifying the window representation. A pseudocode description of the full timeline generation procedure is provided in Appendix~\ref{sec:appendixA}.

\subsection{Operational Context Inference}\label{sec:context}

\paragraph{CLIP embeddings.}\label{par:clip_embedding} \noindent For each window $w_i$, we sample $K$ frames (endpoint excluded) and decode the corresponding keyframes $\{I_{i, 1}, \ldots, I_{i, K}\}$. We use a pretrained \texttt{OpenCLIP} image encoder\footnote{OpenCLIP is an open-source implementation and set of pretrained CLIP models trained on large-scale image--text pairs: \texttt{\url{https://github.com/mlfoundations/open_clip}}} as a fixed feature extractor. Let $\Phi(\cdot)$ denote the frozen image encoder with the pretrained weights specified by our extraction configuration, i.e., the \texttt{(model\_name, pretrained)} pair. For each sampled keyframe $I_{i, k}$ in window $w_i$, we compute an image embedding
\begin{equation}
    \mathbf{z}_{i, k} = \Phi(I_{i, k}) \in \mathbb{R}^{d},
\end{equation}
\noindent and apply $\ell_2$ normalization
\begin{equation}
    \tilde{\mathbf{z}}_{i, k} = \frac{\mathbf{z}_{i, k}}{\|\mathbf{z}_{i, k}\|_2}.
\end{equation}
\tab This normalization has the effect of removing dependence on the magnitude of the embeddings and greatly increasing the stability of the pooled descriptor, $\Pi \in\{\textsc{mean}, \textsc{max}\}$, across sampled frames. To get a single descriptor per window, the normalized keyframe embeddings are then aggregated in a deterministic manner
\begin{equation}
    \mathbf{z}_i = \Pi\big(\{\tilde{\mathbf{z}}_{i, 1}, \ldots, \tilde{\mathbf{z}}_{i, K}\}\big).
\end{equation}
\tab The resulting window descriptors are then stacked up to form a feature matrix $\mathbf{Z}\in\mathbb{R}^{N \times d}$. \\
\paragraph{Context classifier.} \noindent We perform a supervised training on a multiclass logistic regression classifier. This is done using the pooled \texttt{OpenCLIP} window embeddings, $\mathbf{z}_i$. To reduce leakage from temporally adjacent windows within the same incident, we split at the video level. Let $\Labels_{\Context}$ denote the operational context label set defined in Table~\ref{tab:context-defs}. Given a window embedding, $\mathbf{z}_i$, the classifier outputs class probabilities, $\mathbf{p}_i = h_\theta(\mathbf{z}_i)$ giving the predicted probability for each context class $c \in \Labels_{\Context}$
\[
    s_{\text{ctx}, i} = \max_{c} \mathbf{p}_i(c).
\]
\tab If a confidence threshold is enabled, any predictions where $s_{\text{ctx},i} < \tau_{\text{ctx}}$ are labeled as \texttt{LOW\_VIS}.

\subsection{Activity Inference}\label{sec:activity}

\paragraph{Feature representations.} \noindent As an appearance baseline for activity, we compute a per-window representation using the same \texttt{OpenCLIP} encoder and pooling procedure described in~\ref{par:clip_embedding}. For activity, we evaluate two modes: ($1$) \texttt{clip}, which uses only the pooled embedding and ($2$) \texttt{clip\_delta}, which appends three temporal-stability scalars computed from the frame embeddings: the mean and maximum cosine distance between consecutive frame embeddings and the mean per-dimension embedding standard deviation across frames. \\
\tab As a motion-based representation, we compute dense optical flow using Farneb\"ack’s method between consecutive grayscale frames, producing up to $K-1$ flow fields~\cite{farneback2003twoframe}. We summarize motion over the window with a compact, interpretable feature vector
\begin{equation}
    \mathbf{x}_i = \Psi(\{\mathbf{u}_{i, k}\}_{k = 1}^{K - 1}) \in \mathbb{R}^{m}.
\end{equation}
\tab Specifically, for each video, we use the means and standard deviations of the magnitude of flow, magnitude of the direction of flow, magnitude of the global mean flow, simple frame quality measures, and the number of successfully decoded frames $n_{\text{frames}}$ out of $K$ samples from a video. For each video, we then assign $n_{\text{frames}}$ as the number of successfully decoded frames. If less than two frames were decoded, we set the flow-based quantities to zero while retaining $n_{\text{frames}}$. The computed summaries include $12$ dimensional feature vectors. \\
\tab We also examine whether appearance cues help improve activity when combined with motion information. Given $\mathbf{x}^{(\mathrm{clip})}_i \in \mathbb{R}^{d_c}$ and $\mathbf{x}^{(\mathrm{flow})}_i \in \mathbb{R}^{d_f}$, we perform per-block $z$-scoring on the dataset, $b \in \{\textsc{clip}, \textsc{flow}\}$, 
\[
    \tilde{\mathbf{x}}^{(b)}_i = \frac{\mathbf{x}^{(b)}_i - \boldsymbol{\mu}_b}{\boldsymbol{\sigma}_b},
\]
\noindent where $\boldsymbol{\mu}_b$ and $\boldsymbol{\sigma}_b$ are the per-dimension mean and standard deviation of block $b$. The final concatenated feature is then
\begin{equation}
    \mathbf{x}^{(\mathrm{fused})}_i
    =
    \big[\tilde{\mathbf{x}}^{(\mathrm{clip})}_i \,;\, \tilde{\mathbf{x}}^{(\mathrm{flow})}_i\big]
    \in \mathbb{R}^{d_c + d_f}.
\end{equation}
\paragraph{Activity classifier.} \noindent The model is trained to predict over the activity label set $\Labels_{\Act}$ defined in Table~\ref{tab:activity-defs} and is implemented as a multiclass classifier $g_{\varphi}$, trained on window feature vectors $\mathbf{x}_i$. For a window feature vector $\mathbf{x}_i$, $g_{\varphi}$ emits a distribution over class probabilities $\mathbf{q}_i = g_{\varphi}(\mathbf{x}_i)$, where $\mathbf{q}_i(a)$ denotes the predicted probability of activity class $a \in \Labels_{\Act}$. The predicted label with its confidence score is calculated as
\[
    s_{\text{act},i} = \max_{a} \mathbf{q}_i(a).
\]
\tab For uncertain evidence, we support filtering based on a ``confidence'' score. If the score for a given result $s_{\text{act}, i} < \tau_{\text{act}}$, it is classified as \texttt{UNKNOWN}.

% ---------------------------------------------------------------------------------------------------
%     EVALUATION
% ---------------------------------------------------------------------------------------------------

\section{Evaluation}\label{sec:evaluation}

\subsection{Experimental setup}\label{sec:exp_setup}

\noindent We approach the context and activity recognition tasks as supervised window-level classification tasks, by partitioning each video into non-overlapping windows and performing hold-out splits on the video-level. We report the performance on the clean test windows (non-transition regions), and note that transition-flagged windows can potentially have multiple labels, making single-label performance ambiguous. \\ \\
\noindent \textbf{Compared runs.} We present an in-depth study on key aspects of our method. For contextualization, we present a small set of ``easy'' ablations around the frozen visual backbone of the \texttt{OpenCLIP} model, the number of frames we sample within each window, and the pooling method. For activity, we also explore three different representation families: (i) using the motion information provided by optical-flow, (ii) using the embeddings from \texttt{CLIP} with varying frames and pooling and (iii) a fused version of (i) and (ii) where we concatenate the \texttt{CLIP}-derived features with the optical-flow features. In Table~\ref{tab:exp_grid} we provide a complete map of our experiments and how we have split the runs for computing the statistics.
\begin{table*}[h]
    \centering
    \caption{Experiment map for evaluation runs, grouped by task and representation family.}
    \label{tab:exp_grid}
    \footnotesize
    \setlength{\tabcolsep}{5pt}
    \renewcommand{\arraystretch}{1.06}
    \begin{tabular*}{\textwidth}{@{\extracolsep{\fill}} l l l l l @{}}
        \toprule
        \textbf{Task} & \textbf{Run} & \textbf{Model setting} & \textbf{Run} & \textbf{Model setting} \\
        \midrule
        Context (\texttt{CLIP}) & E$1$ & \texttt{ViT-B/32}; $K{=}5$; mean & E$2$ & \texttt{ViT-B/32}; $K{=}10$; mean \\
        Context (\texttt{CLIP}) & E$3$ & \texttt{ViT-L/14}; $K{=}5$; mean & E$4$ & \texttt{ViT-B/32}; $K{=}10$; max \\
        \midrule
        Activity (\texttt{Flow}) & F$1$$^\ast$ & $K{=}5$; $224$$\times$$126$; w$15$; L$3$; it$3$ & F$2$$^\ast$ & $K{=}10$; $320$$\times$$180$; w$15$; L$3$; it$3$ \\
        Activity (\texttt{Flow}) & F$3$$^\ast$ & $K{=}10$; $320$$\times$$180$; w$9$; L$3$; it$3$ & F$4$$^\ast$ & $K{=}10$; $320$$\times$$180$; w$21$; L$3$; it$3$ \\
        Activity (\texttt{Flow}) & F$5$$^\ast$ & $K{=}10$; $320$$\times$$180$; w$21$; L$4$; it$3$ & F$6$$^\ast$ & $K{=}10$; $320$$\times$$180$; w$21$; L$3$; it$5$ \\
        \midrule
        Activity (\texttt{CLIP}) & A$1$$^\ast$ & \texttt{ViT-B/32}; $K{=}10$; mean & A$2$$^\ast$ & \texttt{ViT-B/32}; $K{=}10$; max \\
        Activity (\texttt{CLIP} + $\Delta$) & A$3$$^\ast$ & \texttt{ViT-B/32}; $K{=}10$; mean & A$4$$^\ast$ & \texttt{ViT-B/32}; $K{=}10$; max \\
        Activity (\texttt{CLIP} + $\Delta$) & A$5$$^\ast$ & \texttt{ViT-B/32}; $K{=}5$; mean & A$6$$^\ast$ & \texttt{ViT-B/32}; $K{=}20$; mean \\
        \bottomrule
    \end{tabular*}

    \vspace{0.3em}
    \footnotesize\par $W{\times}H$ denotes resize resolution; $w$ is Farneb\"ack window size; $L$ is the number of pyramid levels; $\textit{it}$ is the number of iterations per level. \\
    $^\ast$ Activity feature configurations eligible for fusion (AF). Only the fused combinations reported in the results section were evaluated.
\end{table*}

\subsection{Results}

\paragraph{Context.} \noindent The results of the window-level context performance are shown in Table~\ref{tab:ctx_results}, which is based on the held-out clean test split. We can see that the performance is consistent across the four different runs, with accuracy ranging from $75.00\%$ to $78.75\%$, and macro-F$1$ ranging from $0.7300$ to $0.7500$. 

\begin{table}[H]
    \centering
    \caption{Context results on held-out test windows.}
    \label{tab:ctx_results}
    \footnotesize
    \setlength{\tabcolsep}{6pt}
    \renewcommand{\arraystretch}{1.08}
    \begin{tabular*}{\columnwidth}{@{\extracolsep{\fill}} l r r @{}}
        \toprule
        \textbf{Run} & \textbf{Accuracy} & \textbf{Macro-F$\textbf{1}$} \\
        \midrule
        E$1$ & $78.75\%$ & $0.7500$ \\
        E$2$ & $78.75\%$ & $0.7300$ \\
        E$3$ & $75.00\%$ & $0.7300$ \\
        E$4$ & $77.50\%$ & $0.7500$ \\
        \bottomrule
    \end{tabular*}
\end{table}

\begin{figure}[h]
    \centering
    \includegraphics[width=\columnwidth]{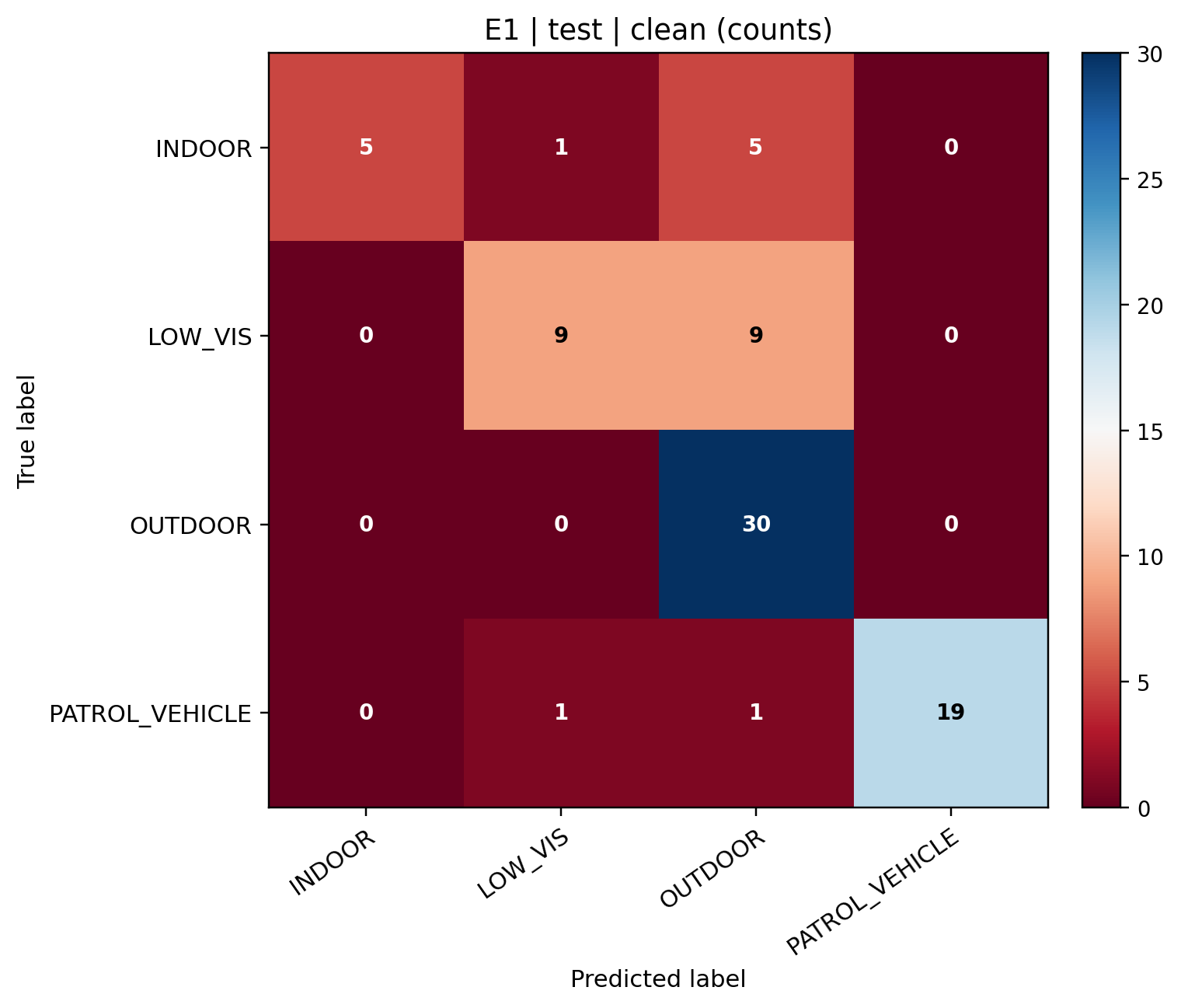}
    \caption{Operational context confusion matrix on the held-out clean test split for E$1$.}
    \label{fig:ctx_cm_grid_1}
\end{figure}

\begin{figure}[h]
    \centering
    \includegraphics[width=\columnwidth]{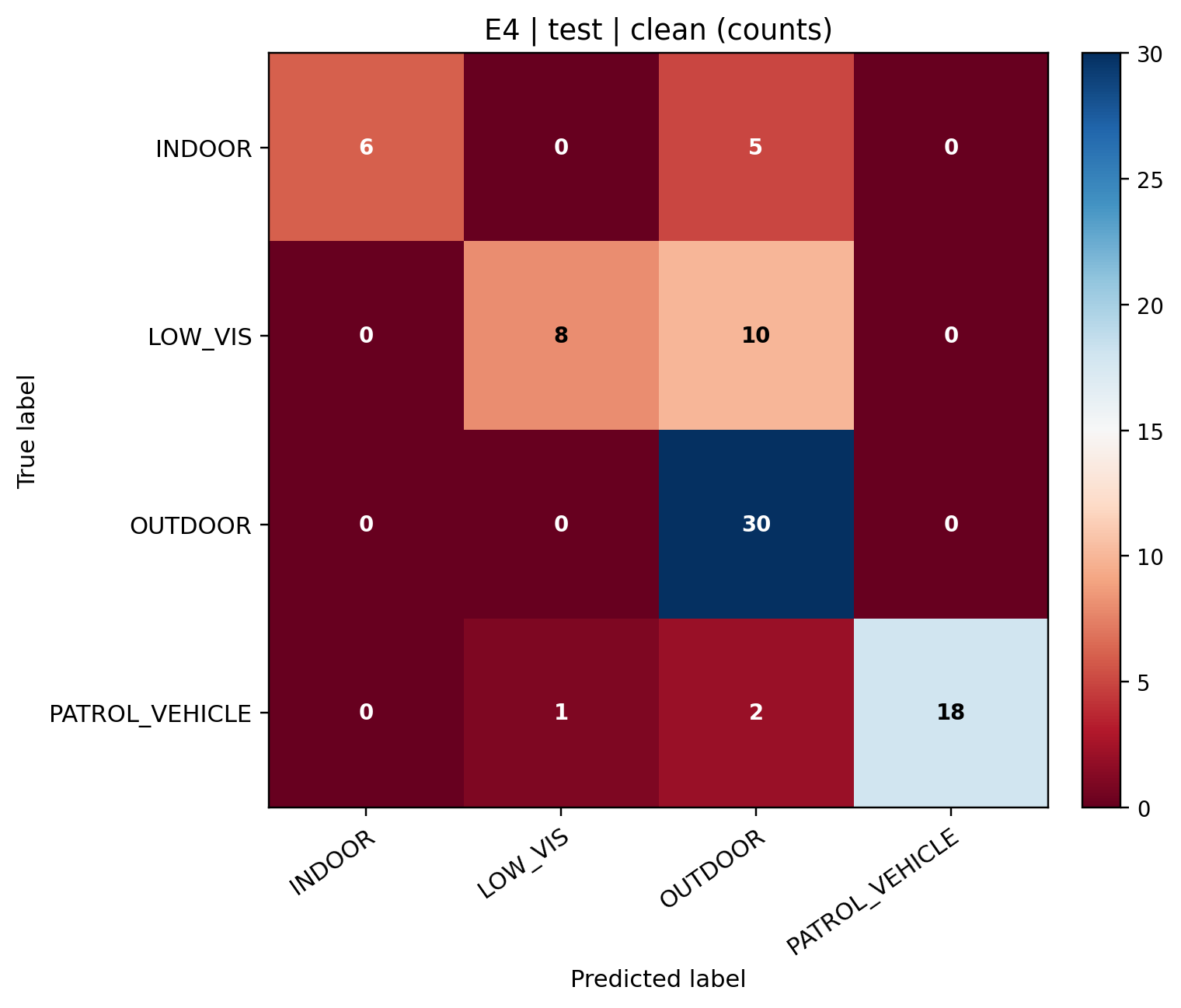}
    \caption[]{Operational context confusion matrix on the held-out clean test split for E$4$.}
    \label{fig:ctx_cm_grid_4}
\end{figure}

\noindent \tab Overall, these results suggest that the context task is quite robust to moderate changes in frame sampling and pooling when the \texttt{CLIP} features are frozen. However, using a larger backbone, such as \texttt{ViT-L/14}, does not seem to provide any benefits in this setting, as E$3$ has the lowest accuracy and does not improve macro-F$1$. Similarly, increasing the number of sampled frames from $5$ to $10$ under mean pooling does not improve performance, as E$2$ has the same accuracy as E$1$ but lower macro-F$1$. On the other hand, max pooling at $K = 10$, as used in E$4$, slightly improves class balance and achieves the best macro-F$1$ score, tied with E$1$. \\
\tab Here we take a closer look at the confusion matrices, displayed in Figures~\ref{fig:ctx_cm_grid_1} and~\ref{fig:ctx_cm_grid_4}. 

\noindent \tab Looking across runs, we see that context errors are not randomly distributed across all the classes. Rather, there are some that are recognized very consistently, such as \texttt{OUTDOOR}, while others such as \texttt{PATROL\_VEHICLE} are recognized relatively consistently, and are usually correctly discriminated from the other contexts. Most errors fall in the boundary cases between \texttt{INDOOR}, \texttt{LOW\_VIS} and \texttt{OUTDOOR}. These errors are not class collapse (i.e., all misclassified as the same class). Rather, they are instances where it is difficult to tell whether a given excerpt of video is one of these three contexts. Therefore, the main challenge in this task is not class collapse, but rather, distinguishing contexts that are very close visually in BWC settings. \\
\tab To help us understand these patterns better, we show the most common recurring confusions across all runs in Appendix~\ref{sec:appendixB}. Figure~\ref{fig:appendix_context_confusions} shows that most of the mistakes happen in a small number of transition types instead of being spread out evenly. \texttt{LOW\_VIS$\rightarrow$OUTDOOR} is the most common mistake in all of the experiments. Confusions involving \texttt{LOW\_VIS} as a target class, on the other hand, are still rare and stable. In general this shows that the main cause of error is not random misclassification, but systematic confusion between contexts that look similar. \\
\paragraph{Activity.} \noindent The detailed activity evaluation reveals a substantially more challenging task than is reflected in the clean-summary metrics. While overall accuracy ranged from $82.50\%$ to $88.33\%$ and macro-F$1$ from $0.3491$ to $0.4535$, performance was largely dominated by the majority class \texttt{ROUTINE} achieving high F$1$. The rarely occurring class \texttt{FOOT\_PURSUIT} was not detected by any run. The main differences between the individual runs are in their F$1$ on the class \texttt{HIGH\_ACTIVITY}, which is the second most frequent class after \texttt{ROUTINE}. The ablations primarily affect performance on the moderately rare dynamic activity class \texttt{HIGH\_ACTIVITY}. \\
\tab Macro-F$1$ is a better evaluation metric than accuracy for this task. The best non-fused run is A$6$, and the best overall run is AF$6$. The fused representations improve the balanced performance over the corresponding \texttt{CLIP}-based models, indicating that appearance and motion are complementary for activity prediction (see Table~\ref{tab:act_results}).
\begin{table}[h]
    \centering
    \caption{Activity results on the held-out test windows.}
    \label{tab:act_results}
    \footnotesize
    \setlength{\tabcolsep}{5pt}
    \renewcommand{\arraystretch}{1.08}
    \begin{tabular*}{\columnwidth}{@{\extracolsep{\fill}} l l r r @{}}
        \toprule
        \textbf{Run} & \textbf{Representation} & \textbf{Accuracy} & \textbf{Macro-F$\textbf{1}$} \\
        \midrule
        A$1$ & \texttt{CLIP} & $85.00\%$ & $0.3740$ \\
        A$2$ & \texttt{CLIP} & $85.83\%$ & $0.3785$ \\
        A$3$ & \texttt{CLIP} + $\Delta$ & $82.50\%$ & $0.3491$ \\
        A$4$ & \texttt{CLIP} + $\Delta$ & $85.00\%$ & $0.3883$ \\
        A$5$ & \texttt{CLIP} + $\Delta$ & $85.00\%$ & $0.3981$ \\
        A$6$ & \texttt{CLIP} + $\Delta$ & $88.33\%$ & $0.4243$ \\
        \midrule
        AF$1$ & \texttt{Fused} & $86.67\%$ & $0.4512$ \\
        AF$2$ & \texttt{Fused} & $86.67\%$ & $0.4058$ \\
        AF$3$ & \texttt{Fused} & $83.33\%$ & $0.3696$ \\
        AF$4$ & \texttt{Fused} & $85.00\%$ & $0.3883$ \\
        AF$5$ & \texttt{Fused} & $85.00\%$ & $0.4334$ \\
        AF$6$ & \texttt{Fused} & $87.50\%$ & $0.4535$ \\
        \bottomrule
    \end{tabular*}
\end{table}

\begin{figure}[h]
    \centering
    \includegraphics[width=\columnwidth]{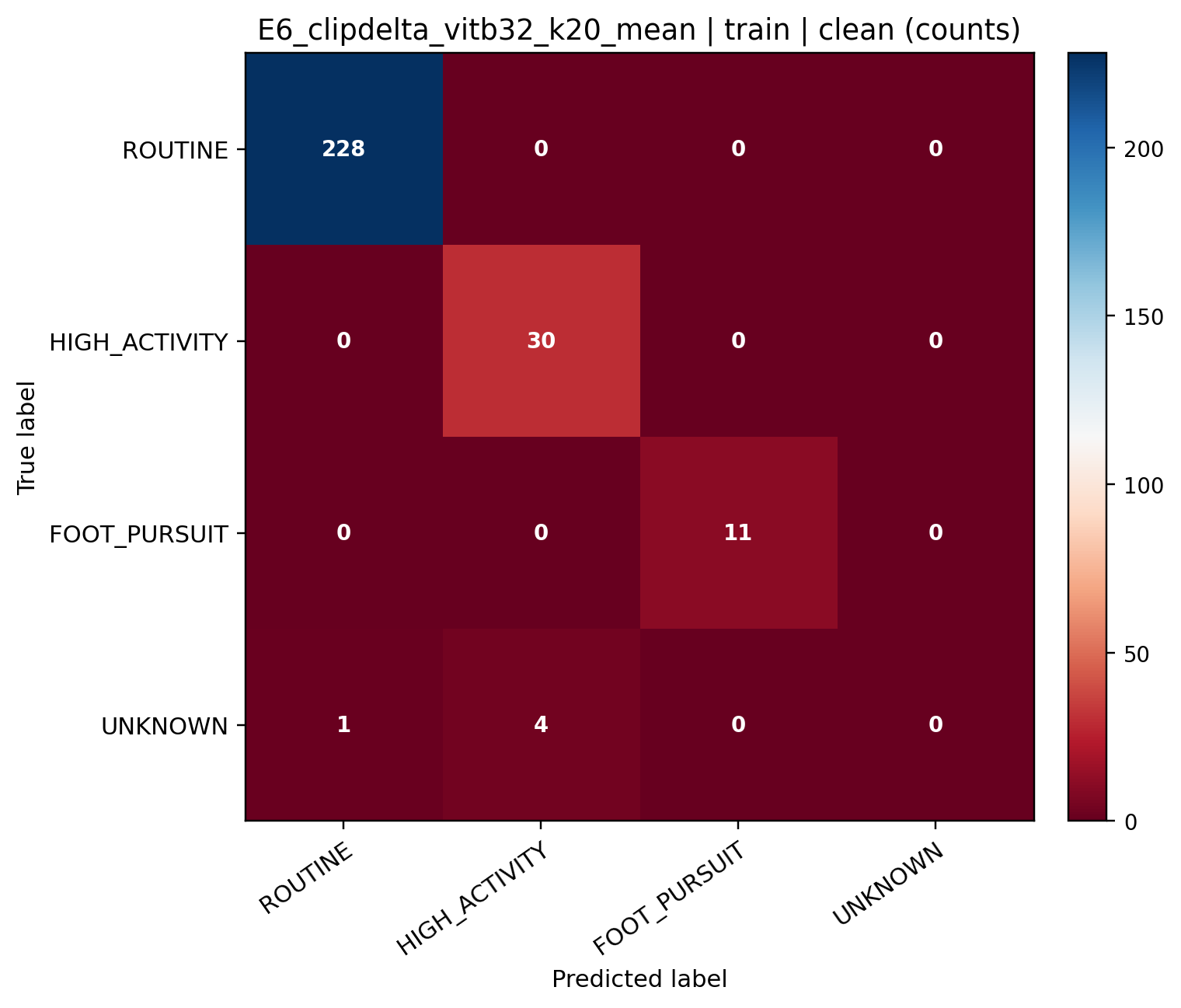}
    
    \vspace{0.5em}
    
    \includegraphics[width=\columnwidth]{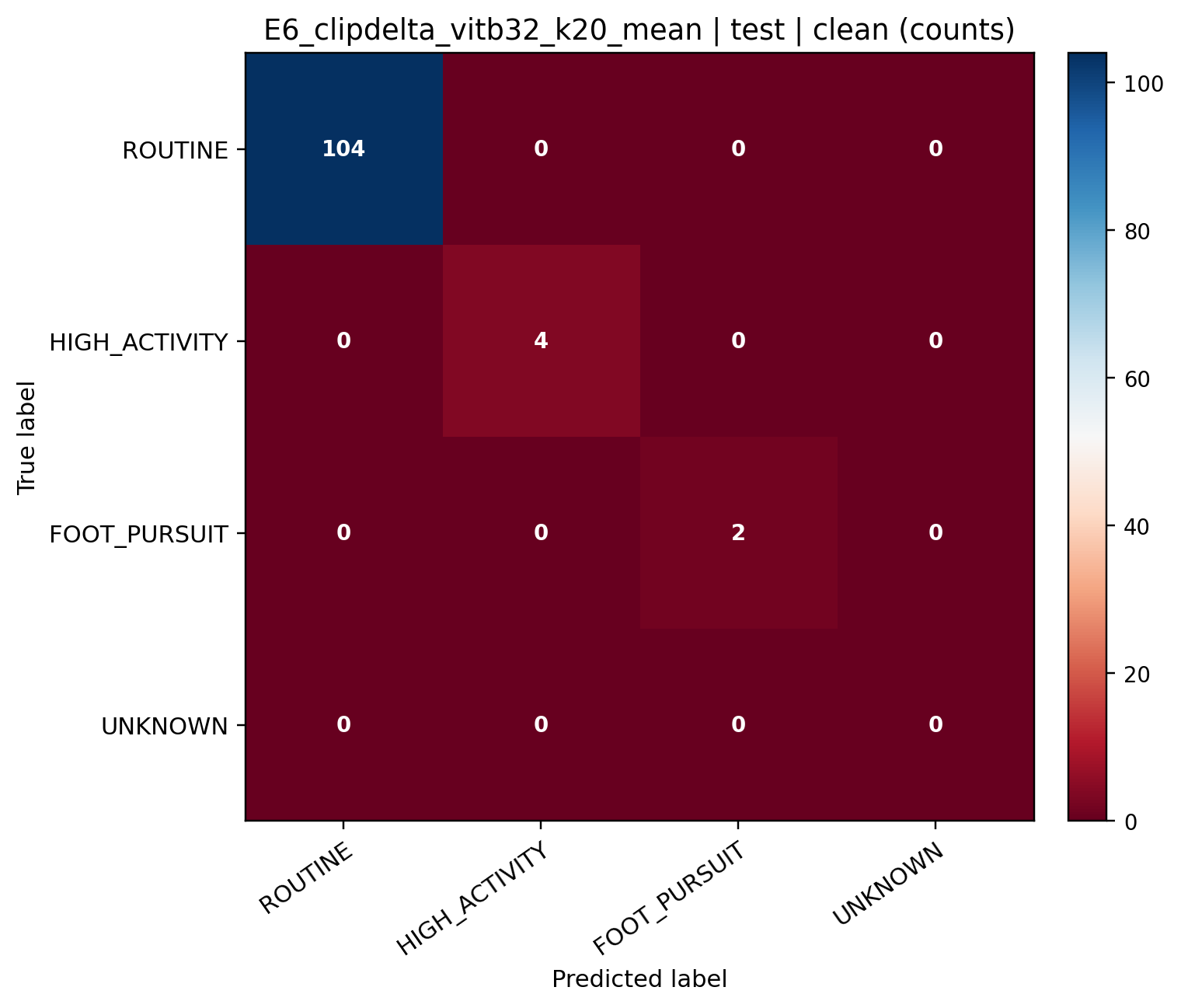}
    \caption{Activity confusion matrices for the strongest non-fused run (A$6$) run. Top: train set. Bottom: test set.}
    \label{fig:act_cm_best}
\end{figure}

\noindent \tab We show the strongest non-fused activity model in Figure~\ref{fig:act_cm_best} on the held-out test set. While the \texttt{ROUTINE} class is recognized very well in both cases, the fused model improves balanced performance most in terms of recognizing more \texttt{HIGH\_ACTIVITY} instances, while \texttt{FOOT\_PURSUIT} remains to be a challenging class in all activity runs.

\noindent \tab Figure~\ref{fig:appendix_activity_clip_confusions} in Appendix~\ref{sec:appendixB} shows the common recurring activity confusions across the \texttt{CLIP}-based runs. The common mistake is \texttt{UNKNOWN} turning into \texttt{HIGH\_ACTIVITY}. This mistake shows up in all models. On the other hand, \texttt{UNKNOWN} turning into \texttt{ROUTINE} does not happen very often. This mistake stays the same every time we run the system. So when the input is not clear the system is more likely to think there is movement than no activity. This shows how difficult it is to tell the difference between weak signals and real dynamic behavior using only appearance features. \\
\tab The fused models in Figure~\ref{fig:appendix_activity_fused_confusions} in Appendix~\ref{sec:appendixB} show a similar pattern. Fusion makes the class balance better overall, but the most common confusion is still \texttt{UNKNOWN$\rightarrow$HIGH\_ACTIVITY}, with a few more counts in some runs. The fact that \texttt{UNKNOWN$\rightarrow$ROUTINE} is the same across all runs shows that the main problem is figuring out how to deal with unclear visual evidence, not how to separate clearly defined activity states. \\
\tab \textbf{RQ$\textbf{4}$} asked if simple baseline models are better at predicting operational context than motion intensity. The findings indicate that the response is affirmative. Even though activity accuracy was sometimes high, this was mostly because of the dominant \texttt{ROUTINE} class. \\
\tab \textbf{RQ$\textbf{5}$} asked if model predictions with low confidence tend to match windows that human annotators also found challenging. The findings indicate that this occurs frequently. For both tasks, uncertainty was mostly found in windows with weak or unclear evidence, such as darkness, occlusion and transition areas. The confidence-based flagging behavior seems to show not only how uncertain the model is, but also how difficult it is for people to label windows.

% ---------------------------------------------------------------------------------------------------
%     DISCUSSION
% ---------------------------------------------------------------------------------------------------

\section{Discussion}\label{sec:discussion}

\noindent The proposed visual timelines suggest that a fixed window approach can provide a useful overview of setting and activity level while remaining consistent with the privacy requirements of the data. The variability seen across the incidents also suggests that a $10$-second window is a useful window size for indexing, model development and review. The approach has several \emph{limitations}: \\
\tab Using fixed-size with non-overlapping windows can cause short transitions near the window boundaries to be ``smoothed over.'' \\
\tab The activity labels use relatively basic optical flow summary features. Although easy to understand the features provide limited descriptive information and may group visually different behaviors under the same label. \\
\tab The size of the labeled set is still limited which determines the complexity of the model.

% ---------------------------------------------------------------------------------------------------
%     CONCLUSION
% ---------------------------------------------------------------------------------------------------

\section{Conclusion}\label{sec:conclusion}

\noindent This work presented a reproducible method for organizing police BWC footage into visual timelines. The approach focuses on human annotation, dataset auditing and baseline models. The results showed that operational context was easier to recover (with F$1$$ = $$ 0.7500$) than motion intensity especially when behavior was unclear (with F$1$$ = $$ 0.4243$). The main contribution is a practical approach for supporting law enforcement review with structured visual information.

% ---------------------------------------------------------------------------------------------------
%     FUTURE WORK
% ---------------------------------------------------------------------------------------------------

\section{Future Work}\label{sec:future_work}

\noindent Next we will focus on linking transcript timestamps to the visual timeline so speech can be reviewed together with changes in scene and activity. We will also add a second review of uncertain windows that uses the surrounding parts of the same incident.

% ---------------------------------------------------------------------------------------------------
%     ACKNOWLEDGMENT
% ---------------------------------------------------------------------------------------------------

\section*{Acknowledgment}

\noindent This research is supported by Grant $15$PBJA-$22$-GG-$03328$-BWCx with the U.S. Department of Justice through its Office of Justice Programs and Bureau of Justice Assistance, awarded to the City of Rochester.

% ---------------------------------------------------------------------------------------------------
%     REFERENCES
% ---------------------------------------------------------------------------------------------------

\bibliographystyle{IEEEtran}
\bibliography{report}

\clearpage
\newpage

% ---------------------------------------------------------------------------------------------------
%     APPENDIX
% ---------------------------------------------------------------------------------------------------

\appendices

\section{Inference and Visual Timeline Construction}\label{sec:appendixA}

\begin{algorithm}[h]
    \caption{\textbf{Visual Timeline Generator}}
    \label{alg:algorithm}
    \small
    \DontPrintSemicolon
    \SetKwInOut{Input}{Input}
    \SetKwInOut{Output}{Output}

    \Input{
    Incident videos $\mathcal{V}$;
    window length $L = 10$;
    frames per window $K$;
    frozen visual encoder $\Phi$;
    pooling operator $\Pi$;
    trained context classifier $f_{\text{ctx}}$;
    activity feature extractor $\Psi$;
    trained activity classifier $f_{\text{act}}$;
    confidence thresholds $\tau_{\text{ctx}}, \tau_{\text{act}}$.
    }

    \Output{
    A mapping of incident videos to window-level timeline records:
    $\{\mathcal{T}(v)\}_{v \in \mathcal{V}}$.
    }

    \ForEach{$v \in \mathcal{V}$}{
        $\mathcal{W} \leftarrow \textbf{MakeWindows}(v, L)$\;
        $\mathcal{T}(v) \leftarrow [\ ]$\;

        \ForEach{$w \in \mathcal{W}$}{
            $F \leftarrow \textbf{SampleFrames}(w, K)$\;

            \tcp{Context pathway}
            $E \leftarrow \{\Phi(x) : x \in F\}$\;
            $\tilde{E} \leftarrow \left\{\frac{e}{\|e\|_2} : e \in E\right\}$\;
            $z_{\text{ctx}} \leftarrow \Pi(\tilde{E})$\;
            $p_{\text{ctx}} \leftarrow f_{\text{ctx}}.\textbf{predict\_prob}(z_{\text{ctx}})$\;
            $\hat{c} \leftarrow \arg\max_{c}\; p_{\text{ctx}}(c)$\;
            $s_{\text{ctx}} \leftarrow \max_{c}\; p_{\text{ctx}}(c)$\;
            \If{$s_{\text{ctx}} < \tau_{\text{ctx}}$}{
                $\hat{c} \leftarrow \texttt{LOW\_VIS}$\;
            }

            \tcp{Activity pathway}
            $x_{\text{act}} \leftarrow \Psi(w, F)$\;
            $p_{\text{act}} \leftarrow f_{\text{act}}.\textbf{predict\_prob}(x_{\text{act}})$\;
            $\hat{a} \leftarrow \arg\max_{a}\; p_{\text{act}}(a)$\;
            $s_{\text{act}} \leftarrow \max_{a}\; p_{\text{act}}(a)$\;
            \If{$s_{\text{act}} < \tau_{\text{act}}$}{
                $\hat{a} \leftarrow \texttt{UNKNOWN}$\;
            }

            Append:
            $\langle
            \texttt{window\_id} = w,\,
            \texttt{start\_time} = \textbf{Start}(w),\,
            \texttt{end\_time} = \textbf{End}(w),\,
            \texttt{context} = \hat{c},\,
            \texttt{context\_score} = s_{\text{ctx}},\,
            \texttt{activity} = \hat{a},\,
            \texttt{activity\_score} = s_{\text{act}}
            \rangle$
            to $\mathcal{T}(v)$\;
        }

        \For{$i \leftarrow 0$ \KwTo $|\mathcal{T}(v)| - 1$}{
            \If{$i = 0$}{
                $\mathcal{T}(v)[i].\texttt{context\_transition} \leftarrow \texttt{True}$\;
                $\mathcal{T}(v)[i].\texttt{activity\_transition} \leftarrow \texttt{True}$\;
            }
            \Else{
                $\mathcal{T}(v)[i].\texttt{context\_transition} \leftarrow
                \big(\mathcal{T}(v)[i].\texttt{context} \neq \mathcal{T}(v)[i - 1].\texttt{context}\big)$\;
                $\mathcal{T}(v)[i].\texttt{activity\_transition} \leftarrow
                \big(\mathcal{T}(v)[i].\texttt{activity} \neq \mathcal{T}(v)[i - 1].\texttt{activity}\big)$\;
            }
        }
    }

    \textbf{Return} $\{\mathcal{T}(v)\}_{v \in \mathcal{V}}$\;
\end{algorithm}

\clearpage
\newpage

\section{Additional Confusion Analysis Figures}\label{sec:appendixB}

\begin{figure}[h]
    \centering
    \includegraphics[width=\linewidth]{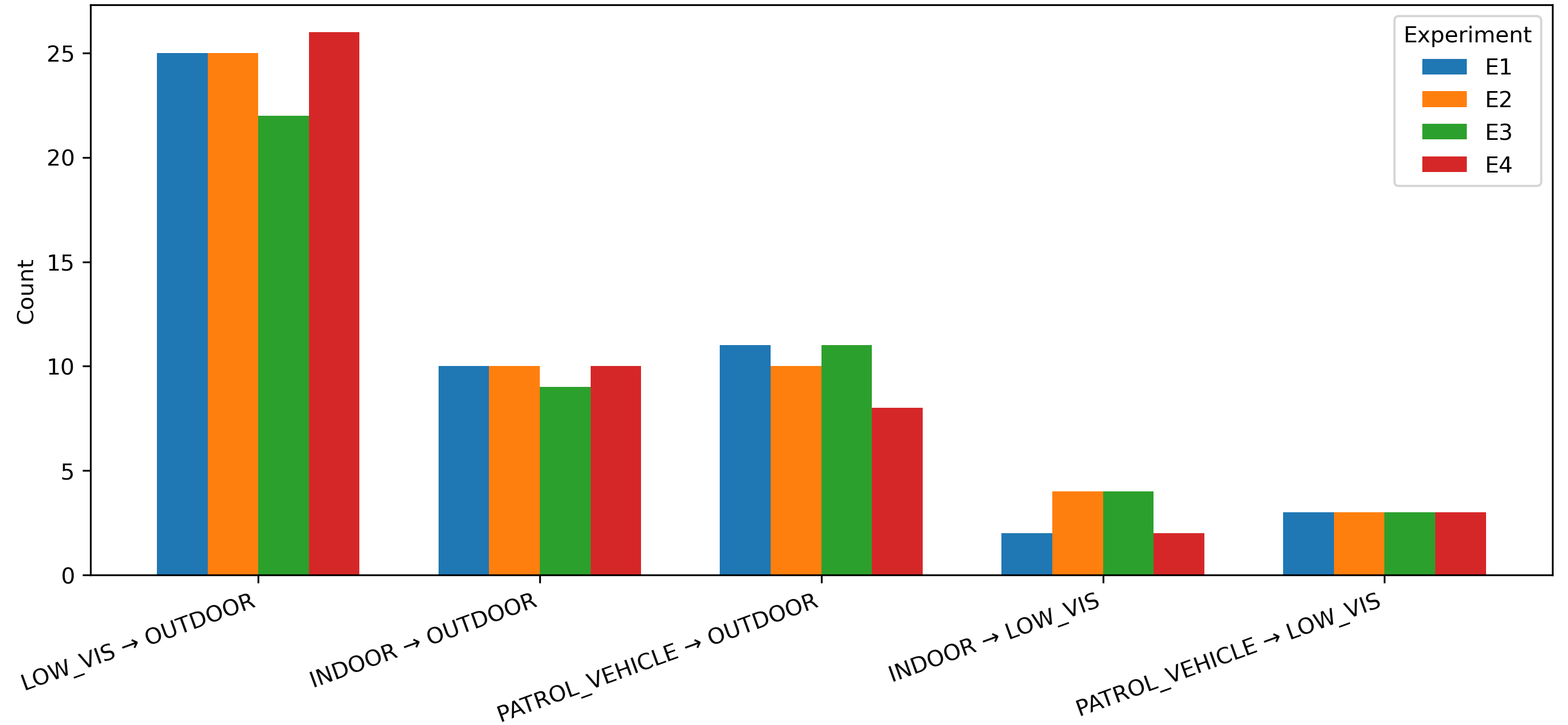}
    \caption{Most frequent recurring context confusions across experiments.}
    \label{fig:appendix_context_confusions}
\end{figure}

\begin{figure}[h]
    \centering
    \includegraphics[width=\linewidth]{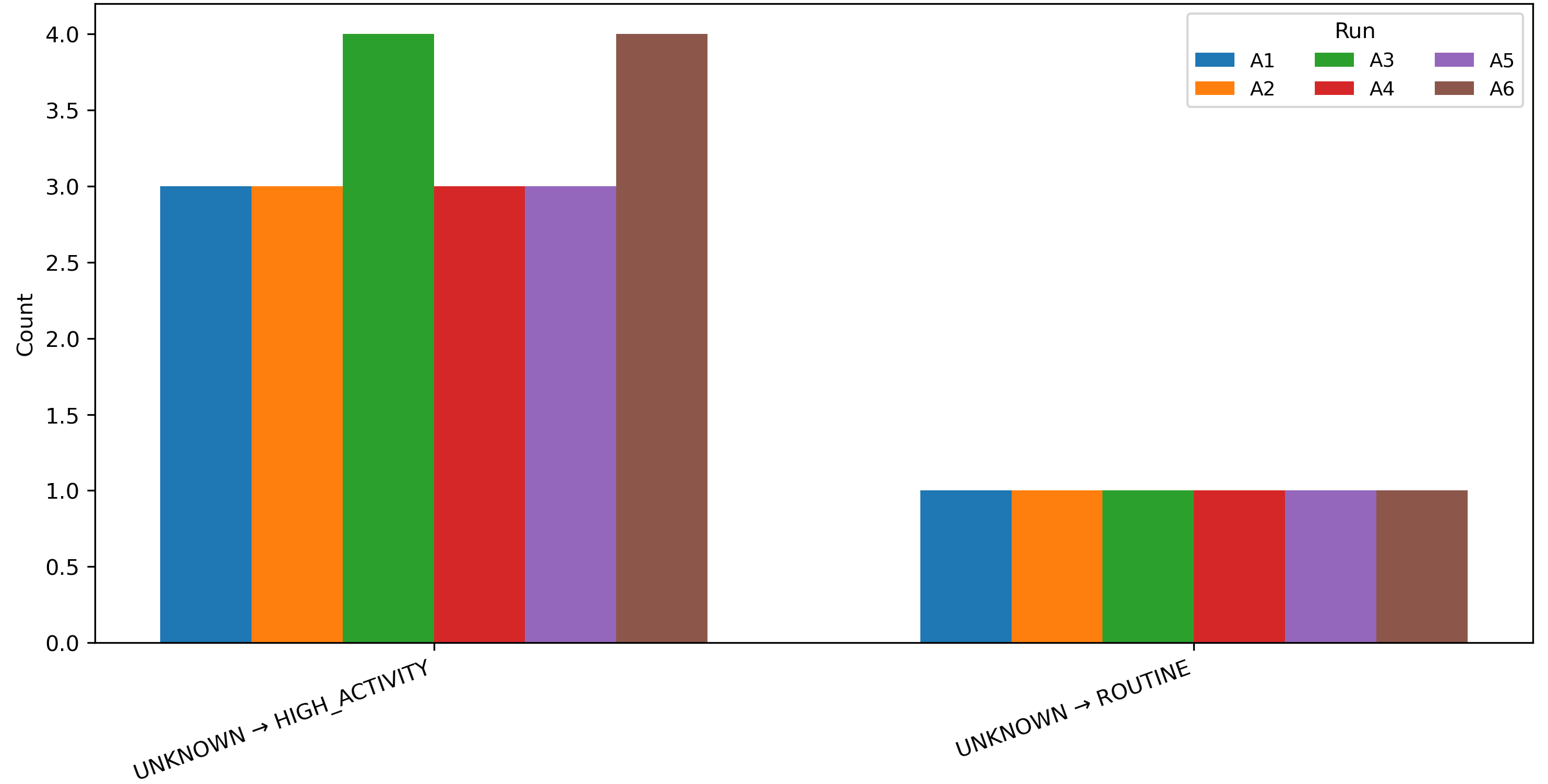}
    \caption{Most frequent recurring activity confusions across \texttt{CLIP}-based runs, shown as counts per confusion type.}
    \label{fig:appendix_activity_clip_confusions}
\end{figure}

\begin{figure}[h]
    \centering
    \includegraphics[width=\linewidth]{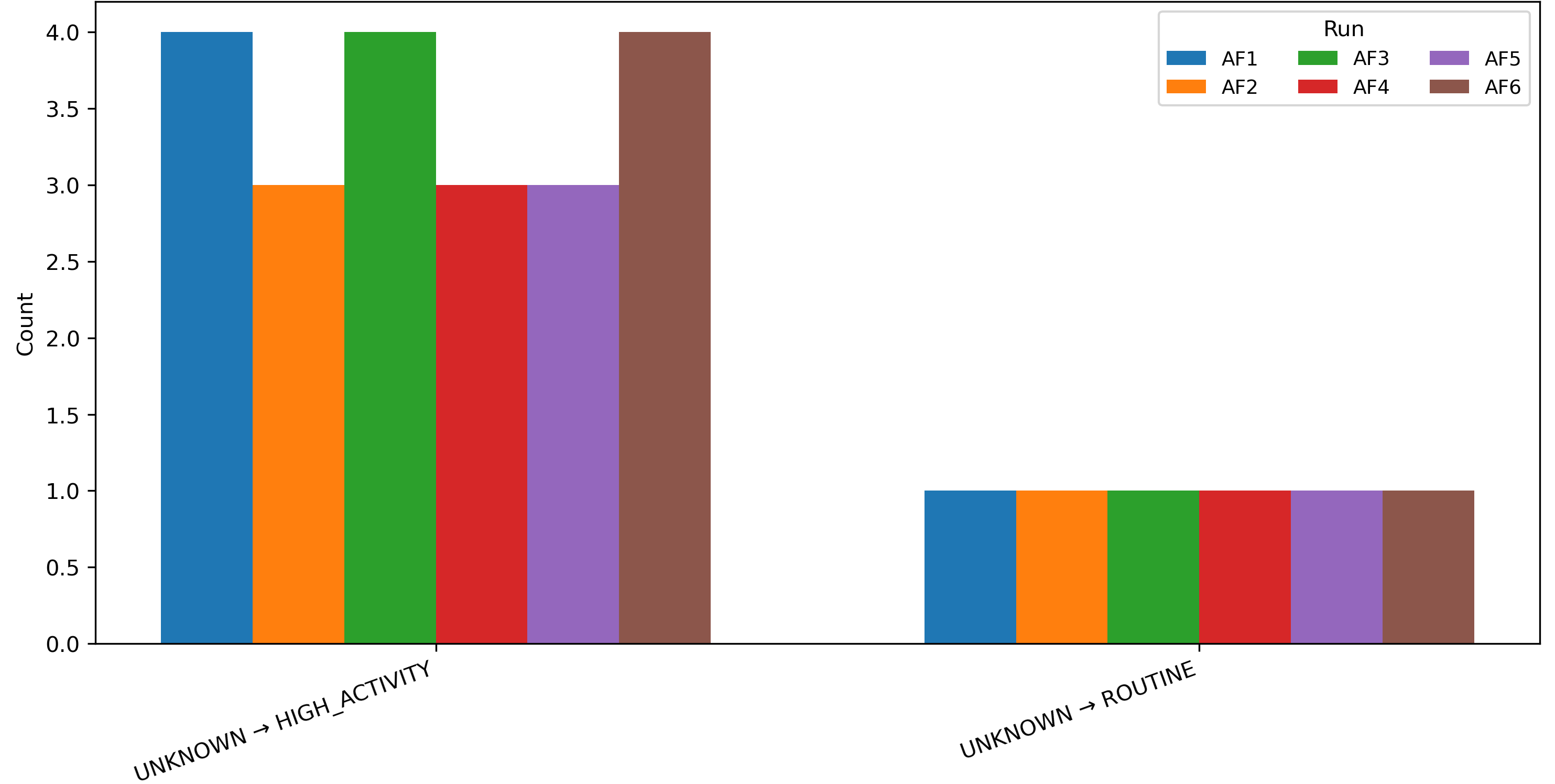}
    \caption{Recurring activity confusions across fused runs, shown as counts per confusion type.}
    \label{fig:appendix_activity_fused_confusions}
\end{figure}

% ---------------------------------------------------------------------------------------------------
%     END DOCUMENT
% ---------------------------------------------------------------------------------------------------

\end{document}